\pdfoutput=1

\documentclass[11pt]{article}

\usepackage[]{ACL2023}
\usepackage{times}
\usepackage{latexsym}
\usepackage{booktabs}
\usepackage{multirow}
\usepackage{listings}
\usepackage{MnSymbol}
\usepackage{makecell}
\usepackage{stfloats}
\usepackage{float}
\usepackage{xcolor}

\def\dataname{COPAL-ID}
\def\categoryculture{Culture}
\def\categoryterm{Local Terminology}
\def\categorylang{Language}

\usepackage[T1]{fontenc}

\usepackage[utf8]{inputenc}

\usepackage{microtype}

\usepackage{inconsolata}

\usepackage{graphicx}
\usepackage{makecell}
\usepackage{pgfplots}
\pgfplotsset{compat=1.17}

\title{COPAL-ID: Indonesian Language Reasoning \\ with Local Culture and Nuances}

\author{
Haryo Akbarianto Wibowo$^1$, Erland Hilman Fuadi$^2$, Made Nindyatama Nityasya$^2$, \\ 
\textbf{Radityo Eko Prasojo$^2$, Alham Fikri Aji$^{1}$} \\
$^1$MBZUAI \quad $^2$Independent Researcher \\
  \texttt{\small haryo.wibowo@mbzuai.ac.ae, \{erland.hilman366,made.nindyatama,radityoeko\}@gmail.com} \\
  \texttt{\small alham.fikri@mbzuai.ac.ae }}

\begin{document}
\maketitle
\begin{abstract}

We present \dataname, a novel, public Indonesian language common sense reasoning dataset. Unlike the previous Indonesian COPA dataset (XCOPA-ID), \dataname{ } incorporates Indonesian local and cultural nuances, and therefore, provides a more natural portrayal of day-to-day causal reasoning within the Indonesian cultural sphere. Professionally written by natives from scratch, \dataname{ } is more fluent and free from awkward phrases, unlike the translated XCOPA-ID. In addition, we present \dataname{ } in both standard Indonesian and in Jakartan Indonesian--a dialect commonly used in daily conversation. \dataname{ } poses a greater challenge for existing open-sourced and closed state-of-the-art multilingual language models, yet is trivially easy for humans. Our findings suggest that general multilingual models struggle to perform well, achieving 66.91\% accuracy on \dataname. South-East Asian-specific models achieve slightly better performance of 73.88\% accuracy. Yet, this number still falls short of near-perfect human performance. This shows that these language models are still way behind in comprehending the local nuances of Indonesian.

\end{abstract}
\section{Introduction}
\begin{table*}[t]
\footnotesize
\centering
\begin{tabular}{@{}p{0.06\linewidth}p{0.21\linewidth}p{0.2\linewidth}p{0.2\linewidth}p{0.2\linewidth}@{}}
\toprule
Category & Premise & Correct Option & Incorrect Option & Note  \\ \midrule
T & Pria itu memperbaharui \textcolor{red}{KK} miliknya \phantom{adawdwad} \textit{(The man updated his \textcolor{red}{KK})}  & Ia baru saja menikah \phantom{asd} \phantom{asdwdhakuahdukwahdukaw} \textit{(He just got married)} & Ia baru saja lulus kuliah \textit{(He had just graduated from college)} & {\textcolor{red}{KK} is a legal document that lists all the family members in a household.} \\

\midrule

L & Rumah tetangga saya baru saja dibobol maling & Dia cuma bisa \textcolor{red}{gigit jari} & Dia meng\textcolor{red}{gigit jari}nya & 
\multirow{2}{\linewidth}{\textit{\textcolor{red}{gigit jari}} is a figure of speech to express helplessness. The 2\textsuperscript{nd} option is more literal.}
\\
& \textit{(My neighbor's house was just broken into by thieves)} & \textit{(He can only bite his fingers)} & \textit{(He bit his finger)} \\
\midrule
T + C &  Anak itu diterima masuk \textcolor{red}{UI} &	Sekeluarga makan \textcolor{red}{nasi kuning} & Sekeluarga makan \textcolor{red}{nasi uduk} 
& \multirow{2}{\linewidth}{\textcolor{red}{UI} is one of the top universities in Indonesia. \textit{\textcolor{red}{Nasi kuning}} is often served for celebrations.} \\ 
& \textit{(That kid was accepted into \textcolor{red}{UI})} & \textit{(The whole family eats \textcolor{red}{yellow rice})} & \textit{(The whole family eats \textcolor{red}{\textit{uduk} rice})} 
\\

\bottomrule
\end{tabular}
\vspace{-0.15cm}
\caption{
\dataname{ } examples. T, L, and C denote \categoryterm, \categorylang, and \categoryculture ~respectively. Some samples may contain multiple categories at once.
}
\vspace{-0.5cm}
\label{tab:examples}
\end{table*}

A predominant challenge in multilingual NLP is to capture the sociolinguistic nuances and contexts that vary from culture to culture \cite{kabra-etal-2023-multi,hershcovich-etal-2022-challenges}. This is especially important in localized language reasoning tasks where knowledge of local context and culture is crucial. For example, while the fact that a ``chanting crowd'' logically follows ``Super Bowl'' might be obvious within the US cultural sphere, the same cannot be said in Japan, where ``Summer Koshien'' is the more locally appropriate context for a ``chanting crowd''. As another example, while ``wearing suits'' follows ``attending a wedding'', within the Indonesian culture ``wearing batik'' is probably the more appropriate consequence. 

Existing multilingual language reasoning datasets such as XNLI \cite{conneau2018xnli} and XCOPA \cite{pontiXCOPAMultilingualDataset2020} do not capture such local nuances because of two reasons. First, they are largely sanitized from localized and cultural elements. General, common-sense instances such as ``water flows'' following ``opening faucet'' are the ones typically found in the datasets.  Second, any cultural element appearing in the dataset is typically based on US/Western context. Even when translated to other languages, it retains the original context while ignoring the cultural mismatch between the context and the common culture of the people speaking the target language. As a consequence, the current language reasoning benchmarks for multilingual models are lacking the crucial aspect of local and cultural context.

To provide a better benchmark for multilingual models that also capture local nuances for Indonesian, we introduce \textbf{\dataname}.\footnote{Data: \url{https://huggingface.co/datasets/haryoaw/COPAL}, Code: \url{https://github.com/haryoa/COPAL-ID}} It follows COPA's~\cite{roemmele2011choice} commonsense causal reasoning format.
\dataname{ }is handcrafted by native long-term Jakarta residents to capture Indonesian cultural and local nuances, especially in Jakarta.
Specifically, we define three categories of locality: \textbf{\categoryculture}, which captures local customs or norm;  \textbf{\categoryterm}, terms commonly known by locals, yet not for outsiders; \textbf{\categorylang}, which tests the nuance of the language, including uses of homonymy and non-compositionality. Each category contains data that can be considered uniquely Indonesian and can be understood as common customs or general knowledge by locals. Additionally, the dataset comes in pairs of two forms: standard Indonesian and colloquial Indonesian~\cite{wibowoIndoCollexTestbedMorphological2021, wibowoSemiSupervisedLowResourceStyle2020a}, with the latter being the go-to form in day-to-day contexts. 

We find that the \dataname{ }dataset is trivially easy for native Jakartans, with our human scorers achieving near-perfect accuracy. However, multilingual NLP models,\footnote{Section~\ref{sec:exp} explains the models that we use for experiments.} both fine-tuned and zero-shot prompted struggle, with many models showing performance close to random chances. In contrast, these models achieve better results in XCOPA-ID. This confirms that although the model might understand the Indonesian language, it struggles to comprehend the cultural aspect that comes with it.

Amongst the models, we find that ChatGPT and GPT-4 perform well, though it is difficult to conclude why given their proprietary nature. Nevertheless, just like the open models, the test on XCOPA-ID yields better results than on our \dataname, denoting the cultural understanding gap.

\section{Related Work}

\paragraph{Multilingual Datasets.} \textbf{XCOPA}~\cite{pontiXCOPAMultilingualDataset2020}, an 11-language dataset translated from the English COPA (\textbf{C}hoice \textbf{o}f \textbf{P}lausible \textbf{A}lternatives)~\cite{roemmele2011choice}, is a commonsense reasoning (CSR) dataset. Each question in the dataset is composed of a premise and two causal alternatives, with the task being to choose the more plausible alternative with respect to the premise. \textbf{XStoryCloze}~\cite{linFewshotLearningMultilingual2022}, a multilingual version of StoryCloze~\cite{mostafazadehCorpusEvaluationFramework2016}, is another CSR dataset that introduces five-sentence stories capturing causal and temporal relations between everyday events in 10 languages. 
XCOPA and XStoryCloze each contain a version in Indonesian through translation, though without much care towards localized nuances and the more natural choices of phrases that locals commonly use. 

Others have built benchmark datasets in Indonesian from the ground up~\cite{mahendra-etal-2021-indonli,leong2023bhasa}, and some have already focused on some cultural aspects, such as proverbs~\cite{liu2023multilingual,kabra-etal-2023-multi} and the cultural nuance within public school exams~\cite{koto2023large}.~\dataname{} differs in its novel focus on cultural commonsense reasoning, which is challenging because it simultaneously tests the cultural knowledge and the logical inference capability of the models.

\vspace{-0.1 cm}

\paragraph{Cultural Aspect.} Some previous work measured the inherent cultural knowledge of LLMs. \citealp{ramezaniKnowledgeCulturalMoral2023}~analyzed whether language models understand cultural norms by evaluating them on two public datasets on morality. Meanwhile, \citealp{dwivedi-etal-2023-eticor} probed LLMs for their knowledge of etiquette norms in five regions. \citealp{linCommonSenseEnglish2021} addressed the challenge of advancing commonsense reasoning beyond English, which is crucial for bridging the gap between different cultures and eliminating language barriers. \citealp{liuVisuallyGroundedReasoning2021} analyzed and created a dataset consisting of image and caption pairs in five languages, including Indonesian, to introduce more languages and cultures to multimodal models.
Similar to these studies, our dataset also enables measurement of LLMs' cultural knowledge, specifically on their cultural commonsense reasoning in Indonesian.

\section{\dataname}
\subsection{Data Language and Demography}
\begin{figure*}[ht!]
  \includegraphics[width=\textwidth]{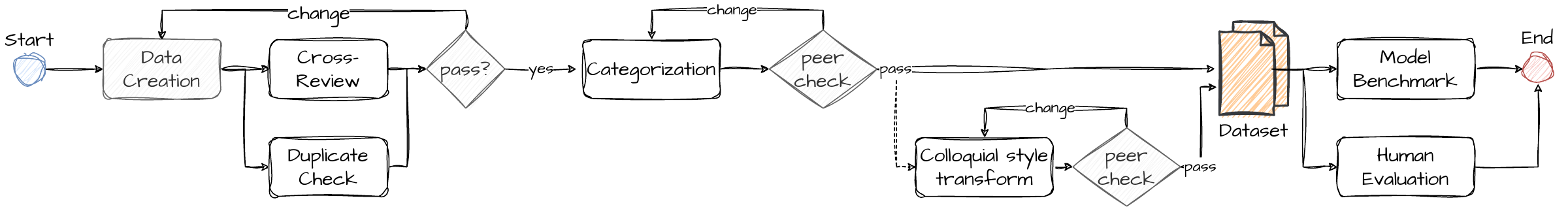}
  \vspace{-0.6cm}
  \caption{\dataname{ }creation and evaluation.}
  \vspace{-0.4cm}
  \label{fig:\dataname}
\end{figure*}
The main purpose of \dataname{ }is to facilitate the benchmarking of NLP models based on their understanding and reasoning capabilities related to Indonesian local and cultural nuances. Relying on existing multilingual benchmarks (e.g., XCOPA~\cite{pontiXCOPAMultilingualDataset2020}, XNLI~\cite{conneau2018xnli}) is inadequate as those data are just translations from English datasets. 

Indonesia itself is rich in culture, consisting of many islands, provinces, and languages. Cultural norms vary in different regions of Indonesia. Therefore, to limit the scope, we focus on the local and cultural aspects of Jakarta, the capital and the most populous city of Indonesia. While the standard Indonesian language is the official language, it's important to note that the Jakartan-Indonesian dialect (often referred to as colloquial Indonesian) is more prevalent in daily conversations in Jakarta than standard Indonesian. Hence, our dataset covers both standard Indonesian and Jakartan-Indonesian dialects, allowing us to evaluate NLP models' proficiency in understanding dialectal variations.

\subsection{Task}

We follow the COPA dataset, where we provide a premise and two plausible alternatives with one that is more likely to happen than the other.

\subsection{Capturing Local Nuances}
\label{sec:\dataname}

We break down local nuances into three categories: culture, local terminology, and languages. Every dataset entry should capture at least one category.

The culture category captures local customs or norms in Indonesia, especially for Jakartan. Local terminology captures long-tail terms or entities that are common and well-known for locals yet alien or long-tail for outsiders. This includes common local foods, animals, places, famous Indonesian public figures, ceremonies, common local abbreviations, and so on. Lastly, language category captures nuances in the language itself, for example, dealing with a figure of speech or word ambiguity. Some examples are shown in Table~\ref{tab:examples}.

\section{Data Creation}
\label{sec:datacreation}

\dataname{ }is created through several steps. As a preliminary step, we formulate the end-to-end plan shown in Figure~\ref{fig:\dataname}. In the data creation step, each of the five data creators (native and raised in Indonesia, accustomed to Jakarta culture) is tasked to create data following the definition in Section~\ref{sec:\dataname}, i.e., to write the premise and both alternatives, in standard Indonesian. Each person is set a target of 110 data (but more is welcome). The resulting data then goes through a check-and-review process. This involves (1) an automated duplicate checker using TF-IDF vectors and (2) a double-blind cross-review process where each created data instance is assigned to two other creators for review. The assignment is distributed uniformly, meaning that there is no bias in the creator-reviewer pairings. The reviewers see the data with the alternatives shuffled, so they do not know the correct answer.

\subsection{Cross Review}

During cross-review, each reviewer performs two tasks for each data instance: (1) to pick the alternative that they deem to be more plausible and (2) to provide a qualitative analysis for the data instance. The analysis concerns several aspects.
\vspace{-0.25cm}
\begin{itemize}
    \setlength\itemsep{-0.3em}
    \item \textbf{Appropriateness}, whether the provided data falls within the definition of a cultural COPA described in Section~\ref{sec:\dataname}. Common concerns here include non-cultural data and out-of-scope/obscure culture.
    \item \textbf{Difficulty}, or lack thereof: the improbable alternative is too obvious.
    \item \textbf{Correctness}, whether the logic, idea, or concept that is relied upon by the data is correct by the common cultural wisdom.
    \item \textbf{Ambiguity}, whether multiple common interpretations can lead to an ambiguity as to which alternative is more plausible.
    \item \textbf{Ethics}, whether the data contains sensitive or discriminating messages.
    \item \textbf{Clarity and format}, whether the provided data has issues with phrasing or spelling.
    \item \textbf{Duplicate}, whether the data reuses the same concept/idea as some other that the reviewer has come across. This is used to supplement the TF-IDF duplicate checker (Section~\ref{sec:dupcheck}).
\end{itemize}
\vspace{-0.25cm}
Data that are answered correctly by both reviewers and have received no qualitative concerns are immediately passed. 
Meanwhile, data that (1) are answered incorrectly by at least one reviewer or (2) have received qualitative concerns are then decided via a discussion by all creators whether to be accepted, rephrased, or sent back for a change.

Changed data go through the same process until at least 550 data are accepted.
Statistics and examples of rejected data can be seen in Appendix~\ref{sec:app-data-creation}.


\begin{table}[]
\small
    \centering
    \begin{tabular}{lrr}
    \toprule
        \multirow{2}{*}{Reasoning Category} & \multicolumn{2}{c}{\#Sample} \\
         & Cause & Effect \\
        \midrule
         Terminology & 186 & 181 \\
         Culture & 136 & 146 \\
         Language & 49 & 57 \\
         \midrule 
         Total & 279 & 280 \\ 
         \bottomrule
    \end{tabular}
    \vspace{-0.2cm}
    \caption{\dataname{} statistic overview}
    \vspace{0.1cm}
    \label{tab:data_statistic}
    
\end{table}

\begin{figure}[]
  \includegraphics[width=200px]{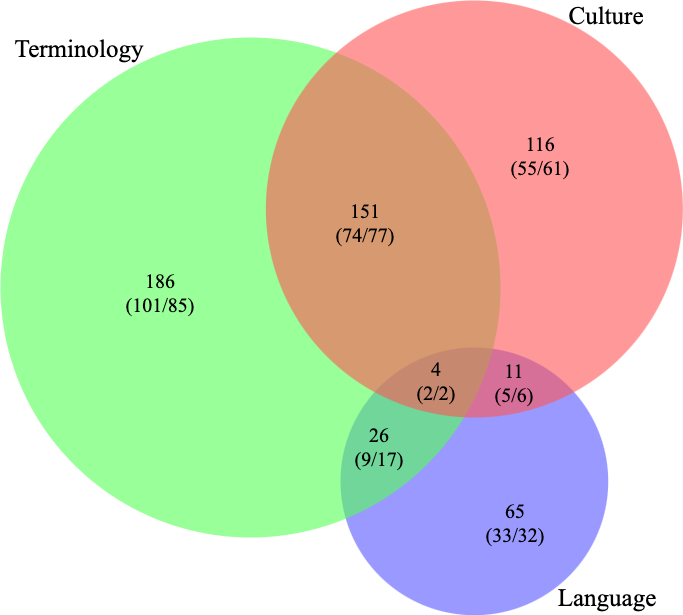}
  \vspace{-0.2cm}
  \caption{\dataname{ }statistic breakdown per-category. The number shows total for (Cause/Effect)}
  \vspace{-0.4cm}
  \label{fig:data_statistic_breakdown}
\end{figure}

\subsection{Duplicate Check}
\label{sec:dupcheck}

We perform a simple duplicate check on the dataset to have more variations of topics or concepts in the dataset.
This duplicate checking was done semi-automatically. First, we automatically identified similar data and grouped them into a cluster. Then, we manually went through each group and checked whether we keep the data or replace it. 

We use a TF-IDF algorithm to get a similarity score between each data. We use stopwords to reduce similarity caused by common words and use both unigram and bigram for the evaluated tokens. We eliminate dissimilar pairs by setting a threshold and then get the final pair pools. 

For the grouping, we first pick a random pair from the pair pools. Then, we check in the pair pools if one of the data has another related pair and then pick it. This process will be done iteratively until we cannot find other associated data in the pair pools. All the related data will be merged into the same group to be evaluated together. From this process, we got 71 groups where each group contains data ranging from 2 to 5. 

The final process is doing a manual check for each group to decide whether to accept or reject the data. For every rejected data, we ask the creator to replace it with a new one. An important thing to note is that for every group, it is possible to accept all the data even though it mentions the same topic, as long as the context is different. Some examples of these can be seen in Appendix~\ref{sec:app-data-creation}, Table~\ref{tab:example_data_duplicate}.

\subsection{Categorization}

We categorize our data to local nuance according to three categories described in Section~\ref{sec:\dataname}. First, the original data creator is asked to annotate the categories of their own data. Then, a different reviewer is asked to validate the label as a peer check. This second reviewer is requested to raise any category label that they deem incorrect and resolve it with the original data creator.

We note that category labeling is very ambiguous for some data. Therefore, one final check is done through a third reviewer. This last reviewer is asked to validate the category label across all \dataname{ }to ensure consistency. Table~\ref{tab:data_statistic} and Figure~\ref{fig:data_statistic_breakdown} show the statistics of \dataname{} categories.

\subsection{Paraphrase to Colloquial Indonesian}
\label{sec:colloquial}
\begin{table*}
\small
    \centering
    \begin{tabular}{ll|p{1.8cm}p{2.0cm}p{1.8cm}p{1.8cm}}
        \toprule
        \textbf{Scenario}  &  & \textbf{Mono (Std.)} & \textbf{Mono (Colloq.)} & \textbf{Cross} & \textbf{Translated} \\
        \midrule
        Finetune   & Training set & ID GEN-X & ID GEN-X  & EN COPA & EN COPA    \\
           & Validation set & XCOPA-ID & XCOPA-ID & EN COPA & EN COPA \\ 
           & Test set &  \dataname & \dataname-C  & \dataname & \dataname-T  \\
           
        \midrule
        In-context & 5-shot examples & \dataname & \dataname-C & EN COPA & EN COPA \\
        & Test set & \dataname  & \dataname-C & \dataname & \dataname-T \\
        \bottomrule
    \end{tabular}
    \vspace{-0.2cm}
    \caption{Data used in finetuning and in-context learning.}
    \vspace{-0.2cm}
    \label{tab:explanation-setup}
\end{table*}
We paraphrase all the datasets into colloquial/Jakartan-dialect. The paraphrase is done by the original data creator while making sure that the meaning of the premise and both alternatives are kept, therefore preserving both the plausibility label and the nuance categorization.

Then, a peer check review is executed to confirm that the newly constructed colloquial dataset still maintains the same meaning, while also making sure that the colloquial text is natural (i.e., commonly used by Jakartan). The data creator will change data entries that do not pass the requirement by the peer reviewer until both are in agreement.

To measure lexical similarity, we compute the BLEU score between the final colloquial dataset and the original standard Indonesian dataset. We obtain a low BLEU score of 3.98, which indicates a high lexical distinction between the two datasets while still keeping the semantics preserved.

\subsection{Human Evaluation}

Lastly, we perform a human evaluation on both standard Indonesian and colloquial Indonesian of \dataname. This evaluation is performed by a completely different group of annotators. We provide the annotators with the premise and two alternatives and ask them to pick the more plausible alternative. Both standard and colloquial Indonesian of \dataname{ }are annotated by two annotators independently. In both datasets, we achieve consistently high accuracy of approximately 95\%, therefore confirming that our dataset is trivial for humans accustomed to Jakartan culture. 

\section{Experiment Setup} \label{sec:exp}

Our data evaluation involves three different setups: monolingual, cross-lingual, translate-test, and colloquial test. Each setup is tailored to a specific scenario. For \textbf{monolingual} setup, we use the same language (Indonesian and Colloquial Indonesian) for training and testing. \textbf{Zero-shot cross-lingual} utilizes the English COPA dataset as the few-shot examples or training data for zero-shot classification and uses the \dataname{ }dataset for testing. In contrast, the \textbf{translate-test} employs an English-translated version of the \dataname{ }for testing instead. We use Seamless-M4T~\cite{communication2023seamless} for translation. 

The evaluation under the monolingual setup is performed twice, one using \dataname{ }with standard Indonesian and the other with the colloquial dataset instead. We forego performing the colloquial testing on the two other setups, noting that the high lexical distinction (Section~\ref{sec:colloquial}) would hamper the translation quality on the translate-test setup and that prior work has noted big degradation in cross-lingual performances of Indonesian local languages even for those with high lexical similarity with standard Indonesian~\cite{winata2023nusax}.

Detailed data and model setup can be seen in Table \ref{tab:explanation-setup} and will be elaborated further in this section.

\begin{table*}[]
\small
\centering

\begin{tabular}{@{ }l|c|cc|c|c}
\toprule
\multirow{2}{*}{\textbf{Test Data}} & & \multicolumn{2}{c|}{Monolingual} & Translate-test & Cross-lingual                                \\  
                                & \textbf{XCOPA-ID} & \textbf{\dataname} & \textbf{\dataname}  & \textbf{\dataname} & \textbf{\dataname}  \\ 
& & (standard) & (colloquial) & (translated) & (standard)\\
                                \midrule
{\textbf{Finetuned models}}  &&&&                                                                       \\
\hspace{8pt}XLMR-Base            &  66.80  &  55.99  &  57.42     &  52.24 & 53.67 \\
\hspace{8pt}XLMR-Large           &  \textbf{79.40}  &  64.22  &  62.97     &  52.06 & 52.95\\
\hspace{8pt}mBERT                &  59.60  &  55.64  &  56.35     &  56.53 & 55.10 \\
\hspace{8pt}IndoBERT             &  67.60  &  61.00  &  60.64     &  - & -\\
\midrule
{\textbf{Open models prompting}} &&&&                                                               \\
\hspace{8pt}Bactrian-X-7B (5-shot)   &  71.20  &  63.51  &  59.39      &  56.35   & 63.69 \\
\hspace{8pt}Llama-2-7B (5-shot)    &  64.20  &  57.96  &  54.29      &  55.81  & 58.86 \\
\hspace{8pt}BLOOMZ-7B (5-shot)            &  76.20  &  66.91  &  58.68     &  56.53  & 65.65\\
\hspace{8pt}PolyLM-13B (5-shot)   &  71.20  &  63.15  &  58.50      & 56.89  & 63.33\\

\midrule
{\textbf{Regional open models prompting}} &&&&   \\  
\hspace{8pt}SeaLLM-7B (5-shot)   &  77.80
 &  73.21  &  64.11    & 59.11  & 68.39\\
\hspace{8pt}Sailor-7B (5-shot)   &  76.80  & \textbf{73.88}  &  \textbf{66.07}      & \textbf{59.57}  & \textbf{73.93}\\
\midrule
{\textbf{Closed models prompting}}  &&&&                                                              \\

\hspace{8pt}ChatGPT (5-shot)      &  90.80  &  76.74  &  76.57  &  -  & -  \\
\hspace{8pt}GPT-4 (5-shot)        &  97.20 &  92.13  &  91.06  &  - & -  \\ \midrule

\textbf{Human}      &  -  &  95.00  &  95.62  &  -   & -  \\
\bottomrule

\end{tabular}
\vspace{-0.2cm}
\caption{Comparison of accuracy score of finetuned, prompting with open and closed models. \textbf{Bold} indicates best results for open models in each column. For IndoBERT, we did not train on COPA-EN as it is inherently not a multilingual model. The score results of \textbf{open models prompting} are the maximum score among different prompt templates and scenarios (few-shots vs zero-shots).}
\vspace{-0.3cm}
\label{main-exp}
\end{table*}

\subsection{Finetuning}

We select four pre-trained models to finetune: MBERT \cite{libovickyHowLanguageNeutralMultilingual2019}, IndoBERT \cite{koto2020indolem}, XLM-R Base, and XLM-R Large \cite{conneauUnsupervisedCrosslingualRepresentation2020}. MBERT \cite{devlin-etal-2019-bert}, XLM-R Base, and XLM-R Large are multilingual models that can handle various languages, including Indonesian language, while IndoBERT is a model that is specifically fine-tuned for Indonesian Language. In this work, we fine-tune our models in monolingual, zero-shot cross-lingual, and translate-test defined above. Due to the unavailability of the training set in XCOPA-ID and \dataname, we use Indonesian COPA data from GEN-X \cite{whitehouseLLMpoweredDataAugmentation2023c}, a multilingual augmented commonsense reasoning dataset produced from GPT-4, instead. Additionally, We use XCOPA-ID as the validation set. Meanwhile, we use English COPA dataset as training and validation set for both the zero-shot cross-lingual and translate-test, and \dataname{ }and \dataname-T as test sets for each scenario, respectively. For each data point, we use two templates to represent the input, which can be found in Appendix~\ref{sec:prompt-template}.

The models are trained with Huggingface's \texttt{transformers}~\cite{wolf2019huggingface}. We do grid-search hyperparameter tuning with batch size of \{8, 16, 32\} and learning rate of \{5$e^{-6}$, 10$e^{-6}$\}. We used a weight decay of 0.01 while the rest of the parameters were set to their default values. We pick the best hyperparameter based on model performance on the validation set. We finetuned each model for 5000 steps, with an early stopping patience of 500. This was done on a 24 GB GPU and completed in under 20 hours in total.

\subsection{Prompting with Language Models}

We test our dataset with in-context learning, which tests the data directly without explicit fine-tuning \cite{brownLanguageModelsAre2020a}, in zero-shot and few-shot settings. For the few-shot setting, for each scenario defined above, we follow the experiment setup defined in MEGA \cite{ahujaMEGAMultilingualEvaluation2023}. We benchmark BLOOMZ~\cite{muennighoff2022crosslingual}, Bactrian-X~\cite{li2023bactrianx}, Llama-2~\cite{touvron2023llama}, and PolyLM~\cite{wei2023polylm} to represent the open-source (multilingual) prompting models. We also include SeaLLM~\cite{seallm} and Sailor~\cite{sailor}, as recent models designed specifically for South-East Asia (SEA) regions. Lastly, we also include ChatGPT\footnote{\url{https://openai.com/blog/chatgpt}} and GPT-4~\cite{openai2023gpt4}\footnote{Both GPT models were accessed in Sep 2023, 4\textsuperscript{th} week} for the proprietary or closed prompting models.

We use five examples for the few-shots scenario with respect to the test data type. In monolingual settings (\dataname{ }and \dataname-C), we use five new in-context examples that we create outside of data produced from section~\ref{sec:datacreation}. We use the English COPA dataset for both cross-lingual and test-translate and test them to \dataname{ }and \dataname-T, respectively. The setup is in Table \ref{tab:explanation-setup}.

To prompt, we benchmark multiple templates since it is widely known that the performance of an LLM depends on its prompt~\cite{liu2023pre}. We use templates from~\cite{ahujaMEGAMultilingualEvaluation2023} (MEGA), BLOOMZ, and LM-Harness.\footnote{\url{github.com/EleutherAI/lm-evaluation-harness}} To check the effect of choosing the language in prompting, we used Indonesian and English templates for each template.\footnote{Template is manually translated to ID if it is only in EN.} After that, we predict the chosen class by computing the logits by applying the template to each choice, comparing each logits, and taking the highest one as the chosen choice. For closed-source models, we use an instruction to make these models output in \texttt{<ANSWER>} format. Then, we extract the predicted answer and match it with the gold label (either $1$ or $2$, which represents the choice that it predicts). Indecisiveness to pick an answer is classified as incorrect instead.
The prompting templates can be found in Appendix~\ref{sec:prompt-template}. 
Prompting on open model is approximately 5 minutes per prompt template in a single A100 GPU.

\begin{figure*}
  \includegraphics[width=\textwidth]{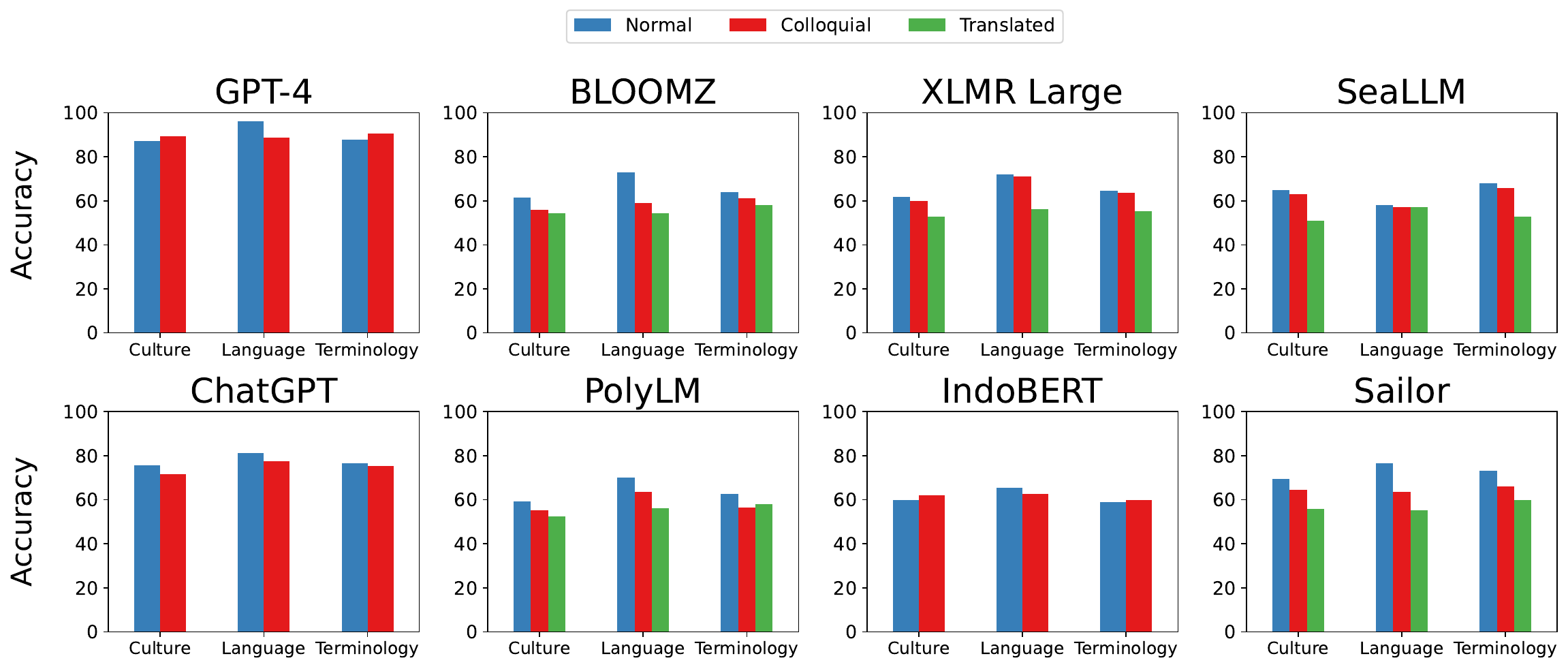}
  \vspace{-0.7cm}
  \caption{Models' accuracy score based on Culture, Language, and Terminology category.}
  \vspace{-0.4cm}
  \label{fig:result-per-category}
\end{figure*}

\section{Experiment Results}

Our experiment results can be seen in Table \ref{main-exp}. By comparing XCOPA-ID and \dataname{ }, it is clear that the latter's accuracy is consistently lower across different models. The best performing open model on XCOPA-ID and \dataname{}, XLMR-Large and BLOOMZ-7B (respectively), have a performance drop of ~15\% and ~9\% (respectively). ChatGPT and GPT-4 also have a performance drop of 14\% and 5\% from XCOPA-ID to COPAL-ID. Smaller performance degradation is seen in SEA-focused models such as SeaLLM and Sailor, signifying that these models have a smaller 'knowledge gap' between general reasoning and locally nuanced reasoning. Yet, Based on this evidence, \dataname{ }can be considered more challenging than XCOPA-ID, which does not incorporate local nuance.

Among open-source models, XLMR-Large exhibits the best performance among the finetuned models, surpassing IndoBERT, which is pre-trained using the Indonesian dataset. On the other hand, for in-context learning, SEA-specific models generally achieve the best performance. For general models, BLOOMZ outperforms others, followed by PolyLM. PolyLM does not surpass BLOOMZ despite having more parameters, showing that it takes other factors, not just a bigger size, for this task. Regarding the closed-source models, GPT-4 outperforms ChatGPT by a significant margin yet cannot beat human performance.

\textbf{Does colloquial Indonesian pose an additional difficulty?} \dataname-C tests the impact of colloquial data on models' performance. It is evident that human performance has comparable results for both colloquial and non-colloquial forms of \dataname, unaffected by it. On the other hand, all of the open models used for prompting performance dropped by about 3-9 percent. Surprisingly, the differences are negligible for each fine-tuned model, demonstrating that finetuning them with the GEN-X dataset is quite robust to colloquial texts. This also indicates a lack of representation of colloquial Indonesian within the pretraining data of these open models. On the other hand, both ChatGPT and GPT-4 models exhibit minimal degradation when evaluated under colloquial Indonesian. However, due to their proprietary nature, it is impossible to pinpoint the reason.

\textbf{For \dataname, does using zero-shot cross-lingual instead of a translate-test improve the score?} Performing translate-test on our dataset tends to degrade the model's performance notably. For instance, XLMR-Large exhibits a significant decline from its standard one by approximately 12\%. We hypothesize that this drop is due to translation errors caused by long-tail words or local terminologies. In this scenario, we recommend using a local-nuance-aware machine translation model to obtain optimal performance, though admittedly such a model (or an open parallel data to build it) currently does not exist.

For finetuned models, the cross-lingual approach (train on EN COPA, then evaluate on \dataname) yields comparably bad performance similar to the translate-test, aside from Sailor. Therefore in this scenario, English data is not really helpful, whether we use it for cross-lingual or translate tests. Interestingly, cross-lingual prompting is visibly better than translate-test, noting that language models can perform in-context learning from a different language.

\begin{table*}
\centering
\small
\begin{tabular}{l|c|cccccccc}
	\toprule
	\multirow{2}{*}{\textbf{Model}}  & \multirow{2}{*}{\textbf{Shots}} & \multicolumn{8}{c}{\textbf{Prompt Template}} \\
	\cmidrule(lr){3-10}
	                                      &   & \textbf{M-ID} & \textbf{M-EN} & \textbf{BL-ID} & \textbf{BL-EN} & \textbf{LH-ID} & \textbf{LH-EN} & \textbf{LLH-ID} & \textbf{LLH-EN} \\
	\midrule
	\multirow{2}{*}{\textbf{Bactrian-X}}  & 5 & 51.16          & 52.24          & 50.45          & 50.27          & 61.90           & 62.43 & \textbf{63.51} & 61.72 \\
	                                      & 0 & 49.19          & 48.84          & 46.51          & 48.84          & 62.08          & 61.18   & \textbf{63.51} & 60.11       \\
	\midrule
	\multirow{2}{*}{\textbf{LLAMA2}}      & 5 & 51.16          & 50.63          & 52.59          & 52.24          & \textbf{57.96}          & \textbf{57.96} & 56.35 & 57.42 \\
                            & 0 & 49.02          & 48.30          & 50.09          & 51.52          & 49.19           & 53.85  & 57.60 & 56.53        \\
	\midrule
	\multirow{2}{*}{\textbf{BLOOMZ}}      & 5 & 54.74          & 55.10  & 60.82          & 57.25          & 65.47          & 65.12    & 65.65 & \textbf{66.91}      \\
	                                      & 0 & 56.89          & 58.68          & 61.90           & 63.15          & 62.79          & 63.86     & 64.22 & 66.55     \\
	\midrule
	\multirow{2}{*}{\textbf{PolyLM (13B)}} & 5 & 54.03          & 55.10           & 54.20           & 53.31          & 62.08          & 62.79       & 62.43 & \textbf{63.15}   \\
	                                      & 0 & 52.77          & 50.45          & 52.06          & 51.16          & 61.90           & 60.47    & 60.64 & \textbf{63.15}      \\
                                       
	\midrule
	\multirow{2}{*}{\textbf{SeaLLM}} & 5 & 69.29          & 67.32           & 70.89           & \textbf{73.21}          & 64.82          & 64.82       & 65.83 & 62.25   \\
	                                      & 0 & 58.04          & 51.96          & 56.61          & 60.54          & 61.79           & 61.96    & 61.54 & 59.21      \\
	\midrule
	\multirow{2}{*}{\textbf{Sailor}} & 5 & 62.68          & 66.43           & 62.50           & 62.86          & 73.57          & 73.57       &  \textbf{73.88} & \textbf{73.88}   \\
	                                      & 0 & 51.61          & 52.14          & 52.86          & 51.96          & 73.57           & 72.68    & 72.81 & 69.95      \\
	\bottomrule
\end{tabular}
\vspace{-0.2cm}
\caption{Indonesian monolingual scenario comparison for every template defined with the addition of juxtaposition between 5-shots and 0-shots. \textbf{Bold} indicates the best results for each model. \textbf{M}=MEGA, \textbf{BL}=BLOOMZ, \textbf{LH}=LM Harness, \textbf{LLH}=Local LM Harness, while \textbf{-ID} and \textbf{-EN} refer to the respective languages.}
\vspace{-0.4cm}
\label{tab:few_prompt}
\end{table*}

\textbf{Do the in-context learning setting and prompt template matter for \dataname?} In Table~\ref{tab:few_prompt}, it is evident that all models benefit from the few-shot scenario on \dataname.\footnote{Also applies on XCOPA with one exception: BLOOMZ, where 0-shot yields best results. (Appendix~\ref{sec:detailed_exp_appendix}, Table~\ref{tab: xcopa_ref_app}).} Moreover, comparing six different templates demonstrates that the best results are yielded with the LM-Harness template in either English or Indonesian languages with comparable scores. This suggests that few-shot scenarios contribute to improving the model performance and underline the requirement of a suitable prompt template for a model to achieve better. As a result, this indicates that those models are sensitive to the choice of prompt template.

\textbf{How is the models' performance across different categories?} Based on Figure \ref{fig:result-per-category}, the Language category of \dataname{ }achieves the best accuracy across all models, while the Culture category performs the worst. This underscores the challenges posed in this category. The Terminology category has a slightly higher overall score than the Culture category. We posit that since the dataset incorporates local nuances that require prior knowledge to answer, it might either conflict with other regions' cultural knowledge or be nonexistent within the embedded knowledge of the pre-training models.

Using colloquial test data also tends to hinder the model performance, especially in the Language category, where the open-source prompting model is more affected. In contrast, fine-tuned models are relatively robust to colloquial text, with some categories demonstrating comparable scores between the original and the colloquial.

\textbf{Would Explicitly Instructing the Models to Reason from a Local Point-of-View Help?} We ran our experiment using previously existing prompts. However, these prompts are in English, and none are explicitly designed for Indonesian causal reasoning. The closest indication for these LLMs to reason based on local aspects is when we translate the prompt into Indonesian.

\begin{table}[]
    \small
    \begin{tabular}{@{}l|ll|ll@{}}
    \toprule
         & \multicolumn{4}{c}{\textbf{Evaluation}} \\
         & \multicolumn{2}{c|}{\textbf{Standard}} & \multicolumn{2}{c}{\textbf{Colloquial}}\\
         \textbf{Prompt Template}  & 0-shot & 5-shot & 0-shot & 5-shot \\
    \midrule
    LM Harness ID & 58.39 & 60.64 & 55.89 & 56.99 \\ 
    \textit{Local} LM Harness ID & 60.23 & 60.59 & 54.96 & 56.48 \\ 
    \midrule
    LM Harness EN & 59.03 & 60.82 & 55.56 & 56.31 \\
    \textit{Local} LM Harness EN & 60.06 & 60.67 & 54.60 & 56.30 \\
    \midrule
    ChatGPT & 67.26 & 76.74 & 62.25 & 74.42 \\ 
    \textit{Local} ChatGPT & 69.94 & 76.74 & 62.79 & 76.57 \\ 
    \midrule
    GPT-4 & 83.00 & 89.80 & 82.28 & 89.80 \\
    \textit{Local} GPT-4 & 86.58 & 92.13 & 85.86 & 91.06 \\
    \bottomrule
    \end{tabular}
    \vspace{-0.2cm}
    \caption{Model performance across prompts with and without local indicator. LM Harness's results are each an average of open model performances (Bactrian-X, Llama-2, BLOOMZ, PolyLM, SeaLLM, Sailor)}
    \vspace{-0.6cm}
    \label{tab:local-indic}
\end{table}

In Table \ref{tab:local-indic}, we observe that when we explicitly prompt the model to reason from the point of view of an "Indonesian accustomed to Jakartan culture", there is a slight increase in overall performance. However, besides GPT-4, the improvement is minimal, and generally, the models still perform poorly, highlighting the challenge of Indonesian cultural reasoning. Prompt details and more comprehensive results are shown in Appendices \ref{sec:prompt-template} and \ref{sec:detailed_exp_appendix}.

\section{Qualitative Analysis}

Upon analyzing the output of the models, we discover some interesting findings where some instances are predicted correctly by at most one model yet answered correctly by all humans. These are displayed in Appendix~\ref{app:qualitativr-analysis}, Table~\ref{tab:examples_error}.

For \textit{nasi uduk} and \textit{nasi kucing} (lit: cat rice), both are dish names with a cultural context where one is often consumed as breakfast while the latter is for dinner. The second example is more intricate. The term \textit{jaga lilin} (translated: keeping the candle), a common black magic practice widely known in Indonesia, requires the model to possess the right cultural knowledge. Other examples show some activities that are not commonly practiced outside Indonesia, such as kissing our parents' hands.

Although the LLMs are trained on multilingual data, some long-tail terms and local nuances are bound to be missed. Without imposing their context explicitly on the models, it would remain challenging for them to infer the correct reasoning.

\section{Conclusion}
We release \dataname{ }to benchmark common sense reasoning with local nuance in Indonesian that consists of culture, terminology, and language in COPA format. We maintain our data quality by having several cross-reviews and automatic duplicate checks. We then benchmark \dataname{ }by evaluating it through in-context learning and finetuning. Our experiment shows that \dataname{ }proves to be challenging for current open-source models across different experiment setups, yet easy for natives. We also provide additional insight that in-context learning is sensitive to the template, and using few-shot examples helps improve the accuracy of the models.

We believe that each region has a different culture, which results in different common sense on some specific things, such as norms and values. Hence, we hope that publishing this dataset encourages the NLP community to build new datasets and models that incorporate diverse local nuances, including Indonesia. 
We leave out how to build a local nuance-aware model for future work. 

\section{Limitations}
Our data is categorized into three categories, which is not granular enough, making more detailed analysis not possible.
Nevertheless, making thoroughly fine-grained categorization is challenging not only in itself but also because we need to increase the size of the dataset as a consequence. After all, the resulting categories would only be useful if an enough number of data instances fall into each of them and therefore are sufficiently statistically representative for analysis purposes.

Additionally, we scope the region only for Jakartan, even though Indonesia, the origin of the Indonesian language, has multiple regions with diverse cultures. Although Jakarta has multi-ethnic citizens and is arguably portrayed the most in mass and social media, not all culture or local nuances are present in Jakarta. Therefore, \dataname{ }has not captured common local nuances in all different Indonesian regions yet. We leave scaling up our dataset to other Indonesia's regions for future work.

\section*{Ethical Considerations}
\dataname{ }has been carefully crafted and reviewed to avoid sensitive and discriminating content. The annotators who are hired for human evaluation are fairly compensated above the minimum wage stipulated by the law in Jakarta.

\bibliography{anthology,custom}
\bibliographystyle{acl_natbib}

\appendix
\newpage

\clearpage
\label{sec:appendix}

\def\ptemp{\texttt{\{p\}}}
\def\cotemp{\texttt{\{c1\}}}
\def\cttemp{\texttt{\{c2\}}}
\def\ttemp{\texttt{\{r\}}}
\def\tltemp{\texttt{\{rl\}}}
\def\newline{\texttt{\textbackslash n}}

\section{Prompt Template}
\label{sec:prompt-template}
This section provides all the templates that we use for prompting and fine-tuning. We first explain the individual variables in each template\footnote{Please note that for the fine-tuning, we treat it as a classification task. Therefore, there is no response in the template.} as follows.
\vspace{-0.3cm}
\begin{itemize}
    \setlength\itemsep{-0.3em}
    \item  \noindent\ptemp: the premise of the data,
    \item  \noindent\cotemp: the first choice of the data,
    \item  \noindent\cttemp: the second choice of the data,
    \item  \noindent\ttemp: the text of the correct choice,
    \item  \noindent\tltemp: lowercased \ttemp
\end{itemize}
\vspace{-0.3cm}
\cotemp{} and \cttemp{} end with a dot (.). No case change is applied unless indicated otherwise.

\vspace{0.3cm}

\hrule

\subsection{MEGA-EN}
\textbf{Prompt} (Cause): \ptemp{.} This happened because...\newline{}Help me pick the more plausible option: - choice1: \cotemp, choice2: \cttemp \newline{}\newline{}\ttemp

\vspace{0.2cm}

\noindent \textbf{Prompt} (Effect): \ptemp{.} As a consequence...\newline{}Help me pick the more plausible option: - choice1: \cotemp, choice2: \cttemp \newline{}\newline{}\ttemp

\vspace{0.3cm}

\hrule

\subsection{MEGA-ID}
\textbf{Prompt} (Cause): \ptemp{.} Ini terjadi karena...\newline{}Bantu saya memilih opsi yang paling mungkin: - opsi1: \cotemp, opsi2: \cttemp \newline{}\newline{}\ttemp

\vspace{0.2cm}

\noindent \textbf{Prompt} (Effect): \ptemp{.} Konsekuensinya...\newline{}Bantu saya memilih opsi yang paling mungkin: - opsi1: \cotemp, opsi2: \cttemp \newline{}\newline{}\ttemp

\vspace{0.3cm}

\hrule

\subsection{Bloomz-EN}
\textbf{Prompt} (Cause): \ptemp{.}\newline{}\newline{}select the most plausible cause:\newline{ }- \cotemp\newline{ }- \cttemp\newline{}\newline{}\ttemp

\vspace{0.2cm}

\noindent \textbf{Prompt} (Effect): \ptemp{.}\newline{}\newline{}select the most plausible effect:\newline{ }- \cotemp\newline{ }- \cttemp\newline{}\newline{}\ttemp

\vspace{0.3cm}

\hrule

\subsection{Bloomz-ID}
\textbf{Prompt} (Cause): \ptemp{.}\newline{}\newline{}pilih penyebab yang paling mungkin:\newline{ }- \cotemp\newline{ }- \cttemp\newline{}\newline{}\ttemp  

\vspace{0.2cm}

\noindent \textbf{Prompt} (Effect): \ptemp{.}\newline{}\newline{}pilih efek yang paling mungkin:\newline{ }- \cotemp\newline{ }- \cttemp\newline{}\newline{}\ttemp

\vspace{0.3cm}

\hrule

\subsection{Fine-tuning}
\textbf{Prompt} (Cause): \ptemp{.}What was the cause?

\vspace{0.1cm}

\noindent\textbf{Prompt} (Effect): \ptemp{.}What happened as a result?

\vspace{0.3cm}

\subsection{LM Harness-EN}

\textbf{Prompt} (Cause): \ptemp{} because \tltemp 

\vspace{0.1cm}

\noindent\textbf{Prompt} (Effect): \ptemp{} therefore \tltemp

\vspace{0.3cm}

\hrule

\subsection{LM Harness-ID}

\textbf{Prompt} (Cause): \ptemp{} karena \tltemp 

\vspace{0.1cm}

\noindent\textbf{Prompt} (Effect): \ptemp{} maka \tltemp

\vspace{0.3cm}

\hrule

\subsection{Local LM Harness-EN}
\textbf{Instruction} : 
Please answer the following question about commonsense causal reasoning from the perspective of someone accustomed to Jakartan culture in Indonesia.

\vspace{0.1cm}

\noindent \textbf{Prompt} (Cause): \ptemp{} because \tltemp 

\vspace{0.1cm}

\noindent\textbf{Prompt} (Effect): \ptemp{} therefore \tltemp

\vspace{0.3cm}

\hrule

\subsection{Local LM Harness-ID}
\textbf{Instruction} : 
Jawablah pertanyaan berikut mengenai penalaran umum sebab akibat dari sudut pandang seseorang yang terbiasa dengan budaya Jakarta di Indonesia.

\vspace{0.1cm}

\noindent \textbf{Prompt} (Cause): \ptemp{} karena \tltemp 

\vspace{0.1cm}

\noindent\textbf{Prompt} (Effect): \ptemp{} maka \tltemp

\vspace{0.3cm}

\hrule

\subsection{ChatGPT and GPT-4}
\textbf{Instruction}: You are an AI assistant whose purpose is to perform open-domain commonsense causal reasoning. You will be provided a premise and two alternatives, where the task is to select the alternative that more plausibly has a causal relation with the premise. Answer as concisely as possible in the same format as the examples below:\footnote{\label{footnote}For zero-shot, this last sentence in the instruction is simply ``Answer as concisely as possible.''}

\vspace{0.1cm}

\noindent \textbf{Prompt} (Cause): \ptemp{.}\newline{}This happened because...\newline{}Help me pick the more plausible option and give me the option index without the text between angle brackets at the end of your answer like this <index>. Be concise. No talk; just go:\newline{} - \cotemp\newline - \cttemp\newline\newline\#\#\# Response:

\vspace{0.1cm}

\noindent \textbf{Prompt} (Effect): \ptemp{.}\newline{}As a consequence...\newline{}Help me pick the more plausible option and give me the option index without the text between angle brackets at the end of your answer like this <index>. Be concise. No talk; just go:\newline{} - \cotemp\newline - \cttemp\newline\newline\#\#\# Response:

\subsection{Local ChatGPT and GPT-4}
\textbf{Instruction}: Please answer the following questions about commonsense causal reasoning from the perspective of someone accustomed to Jakartan culture in Indonesia. You will be provided a premise and two alternatives, where the task is to select the alternative that more plausibly has a causal relation with the premise. Answer as concisely as possible in the same format as the examples below:\footnote{See Footnote~\ref{footnote}.}

\vspace{0.1cm}

\noindent \textbf{Prompt} (Cause): \ptemp{.}\newline{}This happened because...\newline{}Help me pick the more plausible option and give me the option index without the text between angle brackets at the end of your answer like this <index>. Be concise. No talk; just go:\newline{} - \cotemp\newline - \cttemp\newline\newline\#\#\# Response:

\vspace{0.1cm}

\noindent \textbf{Prompt} (Effect): \ptemp{.}\newline{}As a consequence...\newline{}Help me pick the more plausible option and give me the option index without the text between angle brackets at the end of your answer like this <index>. Be concise. No talk; just go:\newline{} - \cotemp\newline - \cttemp\newline\newline\#\#\# Response:

\begin{table*}[bp]
\footnotesize
\centering
\begin{tabular}
{@{}p{0.2\linewidth}p{0.2\linewidth}p{0.2\linewidth}p{0.05\linewidth}p{0.22\linewidth}@{}}
\toprule
Premise & Correct Option & Incorrect Option & Verdict & Note  \\ 

\midrule

    Ia ingin memberbaharui SIM-C miliknya 
    & Ia datang ke kantor polisi
    & Ia datang ke toko ponsel
    & U
    & \multirow{4}{\linewidth}{\textbf{Correctness}. We need to go to SAMSAT (not the police station) to obtain SIM-C (a driving license)} \\
 \textit{(He/she wants to renew his/her SIM-C)} & \textit{(He/she goes to police station)} & \textit{(He/she goes to phone shop)} & & \\

\midrule

    Hari ini ada gangguan sinyal 
    & Perjalanan kereta terlambat
    & Perjalanan pesawat terlambat
    & R
    & \multirow{2}{\linewidth}{\textbf{Appropriateness}. Signalling failures are not uniquely cultural to Indonesia.} \\
 \textit{(Today there is a signalling failure)} & \textit{(Trains are delayed)} & \textit{(Flights are delayed)} & & \\

\midrule

    Dia hendak beribadah pada hari Minggu
    & Dia pergi ke gereja
    & Dia pergi ke wihara
    & R
    & \multirow{2}{\linewidth}{\textbf{Ambiguity}. Churches and Viharas both provide religious services on Sunday.} \\
 \textit{(He/she is going to pray on Sunday)} & \textit{(He/she goes to the church)} & \textit{(He/she goes to the vihara)} & & \\

\midrule

    Saat lebaran, suasana kota Jakarta sangat sepi
    & Masyarakatnya sedang pulang kampung
    & Ada pawai di Jakarta
    & R
    & \multirow{2}{\linewidth}{\textbf{Difficulty}. Parade is never quiet, making this data too easy.} \\
 \textit{(During Eid, Jakarta becomes quiet.)} & \textit{(The citizens are going to their hometown)} & \textit{(There is a parade in Jakarta)} & & \\	

\midrule

    Ayah tidak kunjung tiba karena pesawatnya terlambat berjam-jam
    & Ayah naik <Maskapai A>
    & Ayah naik <Maskapai B>)
    & R
    & \multirow{2}{\linewidth}{\textbf{Ethics}. The original data directly mentions certain airline brands, perpetuating certain stereotypes about them.} \\
 \textit{(Father has not arrived yet because his airline is delayed for hours)} & \textit{(Father flies with <Airline A>)} & \textit{(Father flies with <Airline B>)} & & \\	

\midrule

    Anak itu ketakutan
    & Ia mimpi melihat pocong
    & Ia mimpi melihat porli
    & U
    & \multirow{3}{\linewidth}{\textbf{Clarity}. Pocong is a local ghost, while porli is a typo of Polri (abbreviation of Indonesian national police)} \\
 \textit{(That child is terrified)} & \textit{(He/she dreams of pocong)} & \textit{(He/she dreams of porli)} & & \\
 & {} \\

\bottomrule
\end{tabular}
\vspace{-0.2cm}
\caption{
Candidate of \dataname{} examples from the first iteration that are rejected or need revision. U means ``needs small update/rephrase'', while R means rejected and needs to be replaced with new data.\\ \\ 
}
\label{tab:example_data_cross_review}
\end{table*}

\newpage

\section{Data Creation \& Review}
\label{sec:app-data-creation}
\begin{figure}[h]
  \centering
  \includegraphics[width=7.5cm]{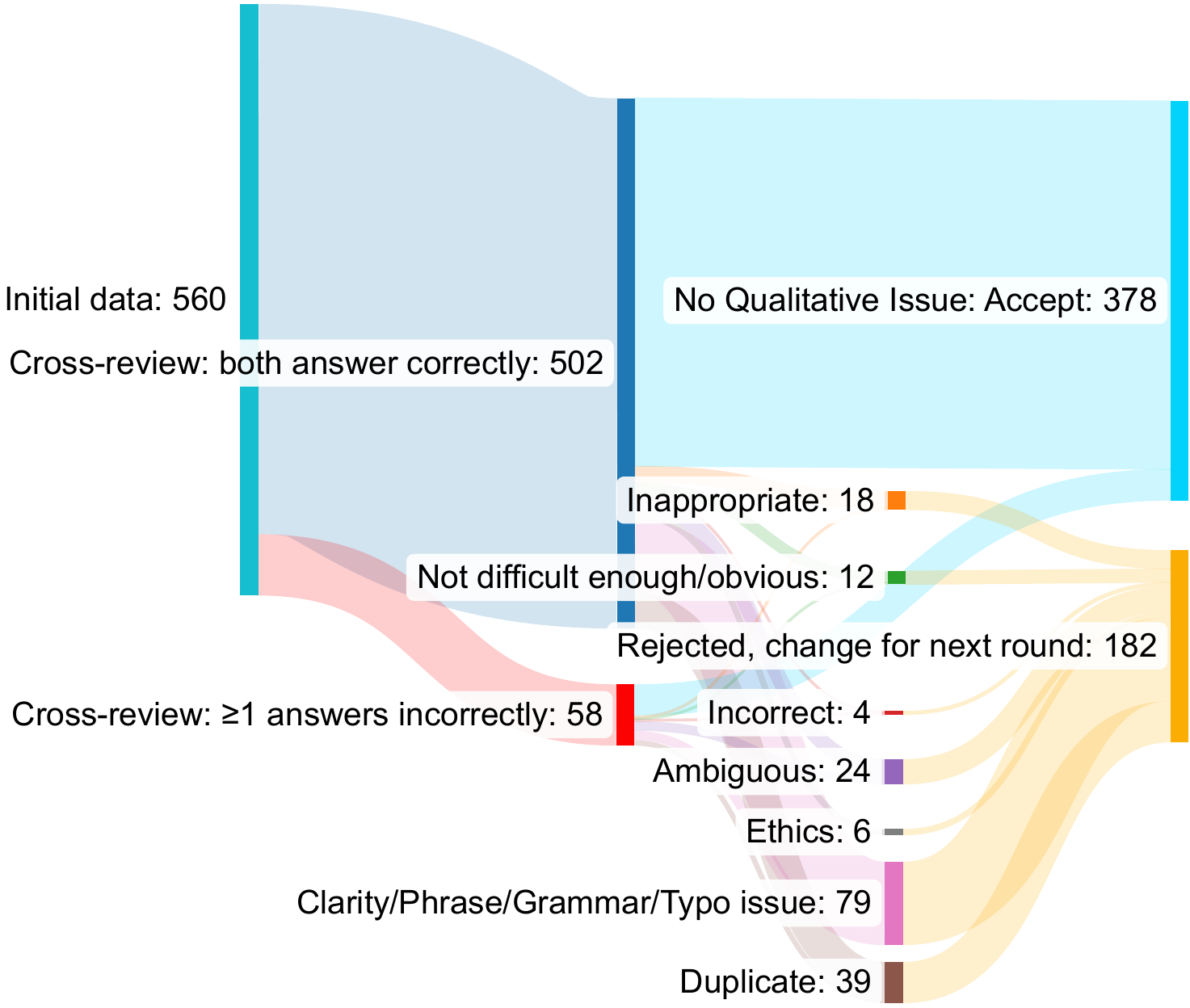}
  \caption{Initial data creation statistics}
  \label{fig:data-creation-sankey}
\end{figure}

Tables~\ref{tab:example_data_cross_review} and \ref{tab:example_data_duplicate} show examples of review decisions. Figure~\ref{fig:data-creation-sankey} shows statistics of the initial review.

\clearpage

\begin{table*}[!ht]
\footnotesize
\centering

\vspace{0.5cm}

\begin{tabular}{@{}p{0.2\linewidth}p{0.2\linewidth}p{0.2\linewidth}p{0.05\linewidth}p{0.22\linewidth}@{}}
\toprule
Premise & Correct Option & Incorrect Option & Verdict & Note  \\ 

\midrule

    Saya mau berangkat sekolah
    & Saya mencium tangan orang tua
    & Saya menunggu bis sekolah
    & A
    & \multirow{4}{\linewidth}{These two data have the same topic of 'kissing parent's hand' and both phrasings are almost identical to each other. In this case, we only accept one of them.} \\
 \textit{(I am going to school)} & \textit{(I kiss my parent's hand)} & \textit{(I wait for the school bus)} & \\
 Anak itu akan berangkat sekolah & Anak itu cium tangan orang tuanya & Anak itu cium dahi orang tuanya & R &  \\ 
 \textit{(That child is going to school)} & \textit{(That child kisses his/her parent's hand)} & \textit{(That child kisses his/her parent's forehead)} & \\ 

\midrule

    Ada kasus demam berdarah di komplek itu
    & Mereka mengadakan penyemprotan
    & Mereka memanggil detektif
    & A
    & \multirow{4}{\linewidth}{Although these two data mention the same topic of ``dengue fever'', the context on how the topic is used is quite different. Thus, we accept both data.} \\
 \textit{(There's a dengue fever case in that area)} & \textit{(They spray the area)} & \textit{(They call a detective)} & \\
 Di musim penghujan, kampungku mengadakan fogging & Pemerintah ingin membasmi demam berdarah & Pemerintah ingin membasmi malaria & A &  \\ 
 \textit{(In rainy season, my village conducts a fogging)} & \textit{(The government wants to prevent dengue fever)} & \textit{(The government wants to prevent malaria)} & \\ 

\bottomrule

\end{tabular}
\caption{
Candidate of \dataname{} examples that are marked as duplicate or similar to each other. A means accepted, R means rejected and needs to be replaced with new data. \\ \\}
\label{tab:example_data_duplicate}

\footnotesize
\centering
\begin{tabular}{@{}p{0.05\linewidth}p{0.2\linewidth}p{0.2\linewidth}p{0.2\linewidth}p{0.22\linewidth}@{}}
\toprule
Category & Premise & Correct Option & Incorrect Option & Note  \\ \midrule

T+C 
    & Malam-malam begini saya kelaparan 
    & Saya keluar berburu \textcolor{red}{nasi kucing} angkringan 
    & Saya keluar berburu \textcolor{red}{nasi uduk} warteg 
    & \multirow{3}{\linewidth}{\textcolor{red}{nasi uduk} and \textcolor{red}{nasi kucing} are dish names, but the former is usually consumed for breakfast while the latter is for late dinner.} \\
 {} & \textit{(I feel hungry at night)} & \textit{(I went out to buy nasi kucing angkringan)} & \textit{(I went out to buy nasi uduk warteg)} \\
 {} \\ 

 \midrule

 C+L
    & Kemarin malam, ia baru selesai \textcolor{red}{jaga lilin}
    & Ia percaya dengan ilmu hitam
    & Ia adalah orang yang taat beribadah 
    & \multirow{3}{\linewidth}{\textcolor{red}{jaga lilin} (keeping the candle alight) is one of the tasks performed in a black magic practice/rituals to get rich.} \\
 {} & \textit{(Yesterday night, he/she just finished \textcolor{red}{jaga lilin})} & \textit{(He/she believed in black magic)} & \textit{(He/she is a religious person)} \\
 {} \\

\midrule

L 
    & Kakek suka \textcolor{red}{sambil menyelam minum air} & Pekerjaan kakek seringkali cepat selesai semuanya & Kakek sering tersedak air & \multirow{2}{\linewidth}{\textcolor{red}{sambil menyelam minum air} is a popular idiom that expresses multitasking ability} \\
 {} & \textit{(Grandfather likes to \textcolor{red}{drink water while diving})} & \textit{(Grandfather often finishes his work quickly)} & \textit{(Grandfather often chokes on water)} & \\

 \midrule

C
    &  Adik yang masih TK akan berangkat ke sekolah & Adik mencium tangan orang tua & Adik menunggu bis sekolah & \multirow{2}{\linewidth}{Kissing the back of parent's hand is a common culture to show respect and love. School buses are rare in Indonesia.}\\ 
 {} & \textit{(Little brother/sister is going to school)} & \textit{(Little brother/sister kisses his/her parent's hand)} & \textit{(Little brother/sister waits for the school bus)} \\

\bottomrule
\end{tabular}
\caption{
\dataname{} examples that is hard for model but trivial for human. T, C, and L depict Terminology, Culture, and Language, respectively.
}
\label{tab:examples_error}

\twocolumn

\end{table*}

\clearpage

\begin{table*}[bp]
\vspace{-2cm}
{\centering
\small
\begin{tabular}{l|l|llllllll}
\toprule
\multirow{2}{*}{\textbf{Model}} &
  \multirow{2}{*}{\textbf{Shots}} &
  \multicolumn{8}{c}{\textbf{Translate-test}} \\ 
 & & 
  \multicolumn{1}{c|}{M-ID} &
  \multicolumn{1}{c|}{M} &
  \multicolumn{1}{c|}{BL-ID} &
  \multicolumn{1}{c|}{BL} &
  \multicolumn{1}{c|}{LH-ID} &
  \multicolumn{1}{c|}{LH-EN} &
  \multicolumn{1}{c|}{LLH-ID} &
  \multicolumn{1}{c}{LLH-EN} \\ 
  \midrule
\multirow{2}{*}{\textbf{Bactrian-X}} &
  5-shots &
  \multicolumn{1}{c|}{50.81} &
  \multicolumn{1}{c|}{50.98} &
  \multicolumn{1}{c|}{47.23} &
  \multicolumn{1}{c|}{47.94} &
  \multicolumn{1}{c|}{54.38} &
  \multicolumn{1}{c|}{53.85} &
  \multicolumn{1}{c|}{56.35} &
  \multicolumn{1}{c}{55.09} \\ 
 &
  0-shot &
  \multicolumn{1}{c|}{50.63} &
  \multicolumn{1}{c|}{47.94} &
  \multicolumn{1}{c|}{50.81} &
  \multicolumn{1}{c|}{49.37} &
  \multicolumn{1}{c|}{54.92} &
  \multicolumn{1}{c|}{53.31}  &
  \multicolumn{1}{c|}{55.99} &
  \multicolumn{1}{c}{54.74} \\ \midrule
\multirow{2}{*}{\textbf{Llama-2}} &
  5-shots &
  \multicolumn{1}{c|}{50.27} &
  \multicolumn{1}{c|}{53.49} &
  \multicolumn{1}{c|}{50.98} &
  \multicolumn{1}{c|}{52.59} &
  \multicolumn{1}{c|}{55.81} &
  \multicolumn{1}{c|}{54.03} &
  \multicolumn{1}{c|}{55.81} &
  \multicolumn{1}{c}{54.91}  \\ 
 &
  0-shot &
  \multicolumn{1}{c|}{50.81} &
  \multicolumn{1}{c|}{49.37} &
  \multicolumn{1}{c|}{49.19} &
  \multicolumn{1}{c|}{51.34} &
  \multicolumn{1}{c|}{55.46} &
  \multicolumn{1}{c|}{54.20} &
  \multicolumn{1}{c|}{56.35} &
  \multicolumn{1}{c}{55.46} \\ \midrule
\multirow{2}{*}{\textbf{Llama-2-Chat}} &
  5-shots &
  \multicolumn{1}{c|}{53.31} &
  \multicolumn{1}{c|}{54.56} &
  \multicolumn{1}{c|}{50.81} &
  \multicolumn{1}{c|}{52.06} &
  \multicolumn{1}{c|}{55.28} &
  \multicolumn{1}{c|}{54.20} &
  \multicolumn{1}{c|}{54.92} &
  \multicolumn{1}{c}{55.10}  \\ 
 &
  0-shot &
  \multicolumn{1}{c|}{50.45} &
  \multicolumn{1}{c|}{48.66} &
  \multicolumn{1}{c|}{52.95} &
  \multicolumn{1}{c|}{50.98} &
  \multicolumn{1}{c|}{53.31} &
  \multicolumn{1}{c|}{54.03} &
  \multicolumn{1}{c|}{55.10} &
  \multicolumn{1}{c}{55.64} \\ \midrule
\multirow{2}{*}{\textbf{BLOOMZ}} &
  5-shots &
  \multicolumn{1}{c|}{53.31} &
  \multicolumn{1}{c|}{53.67} &
  \multicolumn{1}{c|}{54.38} &
  \multicolumn{1}{c|}{53.85} &
  \multicolumn{1}{c|}{55.81} &
  \multicolumn{1}{c|}{54.74}  &
  \multicolumn{1}{c|}{56.53} &
  \multicolumn{1}{c}{54.74} \\ 
 &
  0-shot &
  \multicolumn{1}{c|}{52.95} &
  \multicolumn{1}{c|}{55.46} &
  \multicolumn{1}{c|}{55.10} &
  \multicolumn{1}{c|}{55.81} &
  \multicolumn{1}{c|}{54.92} &
  \multicolumn{1}{c|}{54.20}  &
  \multicolumn{1}{c|}{56.17} &
  \multicolumn{1}{c}{56.17} \\ \midrule
\multirow{2}{*}{\textbf{PolyLM (13B)}} &
  5-shots &
  \multicolumn{1}{c|}{48.12} &
  \multicolumn{1}{c|}{51.34} &
  \multicolumn{1}{c|}{47.23} &
  \multicolumn{1}{c|}{47.58} &
  \multicolumn{1}{c|}{55.99} &
  \multicolumn{1}{c|}{54.20}  &
  \multicolumn{1}{c|}{56.89} &
  \multicolumn{1}{c}{56.53} \\ 
 &
  0-shot &
  \multicolumn{1}{c|}{50.27} &
  \multicolumn{1}{c|}{49.37} &
  \multicolumn{1}{c|}{49.55} &
  \multicolumn{1}{c|}{51.52} &
  \multicolumn{1}{c|}{55.28} &
  \multicolumn{1}{c|}{54.74}  &
  \multicolumn{1}{c|}{55.64} &
  \multicolumn{1}{c}{56.53} \\    \midrule
\multirow{2}{*}{\textbf{SeaLLM}} &
  5-shots &
  \multicolumn{1}{c|}{57.5} &
  \multicolumn{1}{c|}{56.96} &
  \multicolumn{1}{c|}{59.11} &
  \multicolumn{1}{c|}{58.93} &
  \multicolumn{1}{c|}{53.04} &
  \multicolumn{1}{c|}{53.75}  &
  \multicolumn{1}{c|}{55.1} &
  \multicolumn{1}{c}{55.64}  \\
 &
  0-shot &
  \multicolumn{1}{c|}{54.29} &
  \multicolumn{1}{c|}{55.00} &
  \multicolumn{1}{c|}{51.07} &
  \multicolumn{1}{c|}{57.68} &
  \multicolumn{1}{c|}{56.61} &
  \multicolumn{1}{c|}{55.89}  &
  \multicolumn{1}{c|}{55.64} &
  \multicolumn{1}{c}{55.99}  \\ \midrule
  \multirow{2}{*}{\textbf{Sailor}} &
  5-shots &
  \multicolumn{1}{c|}{51.07} &
  \multicolumn{1}{c|}{55.36} &
  \multicolumn{1}{c|}{52.86} &
  \multicolumn{1}{c|}{52.50} &
  \multicolumn{1}{c|}{57.14} &
  \multicolumn{1}{c|}{56.07}  &
  \multicolumn{1}{c|}{57.25} &
  \multicolumn{1}{c}{56.89}  \\
 &
  0-shot &
  \multicolumn{1}{c|}{50.71} &
  \multicolumn{1}{c|}{51.07} &
  \multicolumn{1}{c|}{50.36} &
  \multicolumn{1}{c|}{53.21} &
  \multicolumn{1}{c|}{58.57} &
  \multicolumn{1}{c|}{57.86}  &
  \multicolumn{1}{c|}{59.39} &
  \multicolumn{1}{c}{59.57}  \\ 
  
\bottomrule
\end{tabular}
\vspace{-0.1cm}
\caption{Prompting experiment results for translate-test with respect to model and few shot choices. \textbf{M}=MEGA, \textbf{BL}=BLOOMZ, \textbf{LH}=LM Harness, \textbf{LLH}=Local LM Harness, while \textbf{-ID} and \textbf{-EN} refer to the respective languages.}
\vspace{0.3cm}
\label{tab:translate-test-app}}

{
\small
\centering
\begin{tabular}{l|l|llllllll}
\toprule
\multirow{2}{*}{\textbf{Model}} &
  \multirow{2}{*}{\textbf{Shots}} &
  \multicolumn{8}{c}{\textbf{Colloquial}} \\ 
 & &
  \multicolumn{1}{c|}{M-ID} &
  \multicolumn{1}{c|}{M} &
  \multicolumn{1}{c|}{BL-ID} &
  \multicolumn{1}{c|}{BL} &
  \multicolumn{1}{c|}{LH-ID} &
  \multicolumn{1}{c|}{LH-EN} &
  \multicolumn{1}{c|}{LLH-ID} &
  \multicolumn{1}{c}{LLH-EN} \\ \midrule
\multirow{2}{*}{\textbf{Bactrian-X}} &
  5-shots &
  \multicolumn{1}{c|}{50.27} &
  \multicolumn{1}{c|}{51.88} &
  \multicolumn{1}{c|}{51.52} &
  \multicolumn{1}{c|}{50.27} &
  \multicolumn{1}{c|}{57.96} &
  \multicolumn{1}{c|}{58.04}  &
  \multicolumn{1}{c|}{59.39} &
  \multicolumn{1}{c}{59.21}  \\ 
 &
  0-shot &
  \multicolumn{1}{c|}{48.84} &
  \multicolumn{1}{c|}{51.16} &
  \multicolumn{1}{c|}{49.19} &
  \multicolumn{1}{c|}{48.66} &
  \multicolumn{1}{c|}{56.53} &
  \multicolumn{1}{c|}{57.42}  &
  \multicolumn{1}{c|}{59.21} &
  \multicolumn{1}{c}{56.71}  \\ \midrule
\multirow{2}{*}{\textbf{Llama-2}} &
  5-shots &
  \multicolumn{1}{c|}{50.81} &
  \multicolumn{1}{c|}{52.77} &
  \multicolumn{1}{c|}{50.45} &
  \multicolumn{1}{c|}{49.91} &
  \multicolumn{1}{c|}{54.29} &
  \multicolumn{1}{c|}{53.85}  &
  \multicolumn{1}{c|}{53.49} &
  \multicolumn{1}{c}{53.31}  \\
 &
  0-shot &
  \multicolumn{1}{c|}{51.34} &
  \multicolumn{1}{c|}{50.27} &
  \multicolumn{1}{c|}{51.16} &
  \multicolumn{1}{c|}{51.70} &
  \multicolumn{1}{c|}{52.24} &
  \multicolumn{1}{c|}{52.06}  &
  \multicolumn{1}{c|}{52.06} &
  \multicolumn{1}{c}{52.42}  \\ \midrule
\multirow{2}{*}{\textbf{Llama-2-Chat}} &
  5-shots &
  \multicolumn{1}{c|}{51.70} &
  \multicolumn{1}{c|}{51.16} &
  \multicolumn{1}{c|}{49.91} &
  \multicolumn{1}{c|}{49.19} &
  \multicolumn{1}{c|}{53.67} &
  \multicolumn{1}{c|}{52.59}  &
  \multicolumn{1}{c|}{55.28} &
  \multicolumn{1}{c}{53.49}  \\  
 &
  0-shot &
  \multicolumn{1}{c|}{49.55} &
  \multicolumn{1}{c|}{47.05} &
  \multicolumn{1}{c|}{49.55} &
  \multicolumn{1}{c|}{49.91} &
  \multicolumn{1}{c|}{52.77} &
  \multicolumn{1}{c|}{51.70}  &
  \multicolumn{1}{c|}{52.24} &
  \multicolumn{1}{c}{51.88}  \\ \midrule
\multirow{2}{*}{\textbf{BLOOMZ}} &
  5-shots &
  \multicolumn{1}{c|}{55.28} &
  \multicolumn{1}{c|}{54.92} &
  \multicolumn{1}{c|}{58.50} &
  \multicolumn{1}{c|}{57.96} &
  \multicolumn{1}{c|}{58.14} &
  \multicolumn{1}{c|}{58.68}  &
  \multicolumn{1}{c|}{58.32} &
  \multicolumn{1}{c}{58.50}  \\
 &
  0-shot &
  \multicolumn{1}{c|}{53.31} &
  \multicolumn{1}{c|}{57.78} &
  \multicolumn{1}{c|}{59.39} &
  \multicolumn{1}{c|}{59.57} &
  \multicolumn{1}{c|}{56.17} &
  \multicolumn{1}{c|}{55.10}  &
  \multicolumn{1}{c|}{56.89} &
  \multicolumn{1}{c}{58.50}  \\ \midrule
\multirow{2}{*}{\textbf{PolyLM (13B)}} &
  5-shots &
  \multicolumn{1}{c|}{50.45} &
  \multicolumn{1}{c|}{51.70} &
  \multicolumn{1}{c|}{53.31} &
  \multicolumn{1}{c|}{53.49} &
  \multicolumn{1}{c|}{58.32} &
  \multicolumn{1}{c|}{58.32}  &
  \multicolumn{1}{c|}{58.50} &
  \multicolumn{1}{c}{57.07}  \\ 
 &
  0-shot &
  \multicolumn{1}{c|}{51.16} &
  \multicolumn{1}{c|}{50.09} &
  \multicolumn{1}{c|}{52.42} &
  \multicolumn{1}{c|}{51.52} &
  \multicolumn{1}{c|}{57.07} &
  \multicolumn{1}{c|}{56.71}  &
  \multicolumn{1}{c|}{59.03} &
  \multicolumn{1}{c}{58.32}  \\
  \midrule
\multirow{2}{*}{\textbf{SeaLLM}} &
  5-shots &
\multicolumn{1}{c|}{60.36} & \multicolumn{1}{c|}{62.14} & \multicolumn{1}{c|}{58.39} & \multicolumn{1}{c|}{64.11} & \multicolumn{1}{c|}{56.61} & \multicolumn{1}{c|}{58.04} & \multicolumn{1}{c|}{58.14} & \multicolumn{1}{c}{57.42}  \\
 &
  0-shot &
  \multicolumn{1}{c|}{51.79} & \multicolumn{1}{c|}{50.54} & \multicolumn{1}{c|}{50.00} & \multicolumn{1}{c|}{52.68} & \multicolumn{1}{c|}{56.07} & \multicolumn{1}{c|}{56.25} & \multicolumn{1}{c|}{56.17} & \multicolumn{1}{c}{54.56}  \\ \midrule
\multirow{2}{*}{\textbf{Sailor}} &
  5-shots &
 \multicolumn{1}{c|}{53.21} & \multicolumn{1}{c|}{54.11} & \multicolumn{1}{c|}{55.89} & \multicolumn{1}{c|}{55.36} & \multicolumn{1}{c|}{65.00} & \multicolumn{1}{c|}{63.93} & \multicolumn{1}{c|}{65.12} & \multicolumn{1}{c}{64.58}  \\ 
 &
  0-shot &
  \multicolumn{1}{c|}{48.75} & \multicolumn{1}{c|}{49.11} & \multicolumn{1}{c|}{51.25} & \multicolumn{1}{c|}{51.25} & \multicolumn{1}{c|}{66.07} & \multicolumn{1}{c|}{64.64} & \multicolumn{1}{c|}{63.51} & \multicolumn{1}{c}{62.79}  \\
\bottomrule
\end{tabular}
\vspace{-0.2cm}
\caption{Colloquial experiment results. \textbf{M}=MEGA, \textbf{BL}=BLOOMZ, \textbf{LH}=LM Harness, \textbf{LLH}=Local LM Harness, while \textbf{-ID} and \textbf{-EN} refer to the respective languages.}
\label{tab:collo_xcopas_app}
}
\vspace{2cm}
\end{table*}

\section{Qualitative Analysis}
\label{app:qualitativr-analysis}
Table \ref{tab:examples_error} provides examples of qualitative result analysis that showcase how our tested models struggle to understand the cultural nuance in \dataname{}.

\section{Detailed Experiment Results} \label{sec:detailed_exp_appendix}

Tables \ref{tab:translate-test-app}, \ref{tab:collo_xcopas_app}, \ref{tab:crosslingual_app}, and \ref{tab: xcopa_ref_app} provide prompting results for cross monolingual \& translate-test, colloquial, cross-lingual, and XCOPA respectively.

\newpage

\section{Answer Rejection From Closed LMs}

Table \ref{tab:reject} shows the number of instances for which closed-source LMs reject to answer.

\begin{table}[!h]
\centering
\normalsize
\begin{tabular}{cc}
\toprule
  Normal   & Colloquial \\
\midrule
   15  &  11 \\
\bottomrule
\end{tabular}
\caption{Number of instances GPT-4 refused to answer. ChatGPT answered all questions. }
\label{tab:reject}
\end{table}

\clearpage

\begin{table*}[!ht]

{
\centering
\small
\begin{tabular}{l|l|l|l|l|l|l|l|l}
\toprule
\textbf{Model}        & \textbf{M-ID} & \textbf{M}  & \textbf{BL-ID} & \textbf{BL} & \textbf{LH-ID} & \textbf{LH-EN} & \textbf{LLH-ID} & \textbf{LLH-EN} \\\midrule
\textbf{Bactrian-X}   & 51.52   & 49.73 & 50.45     & 50.81  & 60.29  & 59.93 &  63.69 & 60.64      \\ 
\textbf{Llama-2}       & 51.70  & 51.34 & 50.45     & 50.45 & 56.89  & 55.81  & 58.86 & 56.71    \\ 
\textbf{Llama-2-Chat}  & 53.13   & 52.59  & 50.09        & 49.91 & 55.99  & 55.28 & 55.46 & 56.35      \\ 
\textbf{BLOOMZ}       & 55.10   & 54.56 & 55.64     & 57.60  & 62.97  & 62.97 &  65.65 & 65.29   \\ 
\textbf{PolyLM (13B)} & 53.67   & 51.16 & 55.64    & 51.52  & 63.33  & 61.90 & 62.97 & 62.97    \\ 
\textbf{SeaLLM} & 68.39 & 65.71 & 68.04 & 68.04 & 61.79 & 62.68 & 63.51 & 62.61    \\ 
\textbf{Sailor} & 62.50 & 67.68 & 60.54 & 59.64 & 73.93  & 71.43 & 71.43 & 72.09    \\ 
\bottomrule
\end{tabular}
\vspace{-0.2cm}
\caption{Cross-lingual experiment results (5-shot).}
\vspace{0.3cm}
\label{tab:crosslingual_app}
}

{
\centering
\small
\begin{tabular}{l|l|llllll}
\toprule
\multirow{2}{*}{\textbf{Model}} &
  \multirow{2}{*}{\textbf{Shots}} &
  \multicolumn{6}{c}{\textbf{XCOPA}}  \\ 
 & &
  \multicolumn{1}{c|}{M-ID} &
  \multicolumn{1}{c|}{M} &
  \multicolumn{1}{c|}{BL-ID} &
  \multicolumn{1}{c|}{BL} &
  \multicolumn{1}{c|}{LH-ID} &
  \multicolumn{1}{c}{LH-EN}   \\ \midrule
\multirow{2}{*}{\textbf{Bactrian-X}} &
  5-shots &
  \multicolumn{1}{l|}{52.60} &
  \multicolumn{1}{l|}{51.60} &
  \multicolumn{1}{l|}{55.00} &
  \multicolumn{1}{l|}{51.00} &
  \multicolumn{1}{l|}{69.60} &
  \multicolumn{1}{l}{71.20}  \\ 
 &
  0-shot &
  \multicolumn{1}{l|}{54.20} &
  \multicolumn{1}{l|}{53.00} &
  \multicolumn{1}{l|}{53.00} &
  \multicolumn{1}{l|}{54.80} &
  \multicolumn{1}{l|}{67.20} &
  \multicolumn{1}{l}{67.60}    \\ \midrule
\multirow{2}{*}{\textbf{Llama-2}} &
  5-shots &
  \multicolumn{1}{l|}{56.00} &
  \multicolumn{1}{l|}{56.40} &
  \multicolumn{1}{l|}{56.80} &
  \multicolumn{1}{l|}{58.80} &
  \multicolumn{1}{l|}{63.20} &
  \multicolumn{1}{l}{64.20}    \\
 &
  0-shot &
  \multicolumn{1}{l|}{51.00} &
  \multicolumn{1}{l|}{50.00} &
  \multicolumn{1}{l|}{51.60} &
  \multicolumn{1}{l|}{51.80} &
  \multicolumn{1}{l|}{61.00} &
  \multicolumn{1}{l}{60.60}    \\ \midrule
\multirow{2}{*}{\textbf{Llama-2-Chat}} &
  5-shots &
  \multicolumn{1}{l|}{54.80} &
  \multicolumn{1}{l|}{59.60} &
  \multicolumn{1}{l|}{57.00} &
  \multicolumn{1}{l|}{60.80} &
  \multicolumn{1}{l|}{60.20} &
  \multicolumn{1}{l}{61.80}    \\  
 &
  0-shot &
  \multicolumn{1}{l|}{51.60} &
  \multicolumn{1}{l|}{51.00} &
  \multicolumn{1}{l|}{51.00} &
  \multicolumn{1}{l|}{49.20} &
  \multicolumn{1}{l|}{58.80} &
  \multicolumn{1}{l}{57.40}   \\ \midrule
\multirow{2}{*}{\textbf{BLOOMZ}} &
  5-shots &
  \multicolumn{1}{l|}{61.60} &
  \multicolumn{1}{l|}{62.00} &
  \multicolumn{1}{l|}{66.60} &
  \multicolumn{1}{l|}{71.20} &
  \multicolumn{1}{l|}{71.80} &
  \multicolumn{1}{l}{69.60}    \\
 &
  0-shot &
  \multicolumn{1}{l|}{59.60} &
  \multicolumn{1}{l|}{71.00} &
  \multicolumn{1}{l|}{66.80} &
  \multicolumn{1}{l|}{76.20} &
  \multicolumn{1}{l|}{60.60} &
  \multicolumn{1}{l}{59.80} \\ \midrule
\multirow{2}{*}{\textbf{PolyLM (13B)}} &
  5-shots &
  \multicolumn{1}{l|}{48.60} &
  \multicolumn{1}{l|}{49.00} &
  \multicolumn{1}{l|}{48.80} &
  \multicolumn{1}{l|}{50.00} &
  \multicolumn{1}{l|}{71.20} &
  \multicolumn{1}{l}{68.60}  \\ 
 &
  0-shot &
  \multicolumn{1}{l|}{47.00} &
  \multicolumn{1}{l|}{49.60} &
  \multicolumn{1}{l|}{48.60} &
  \multicolumn{1}{l|}{50.20} &
  \multicolumn{1}{l|}{70.20} &
  \multicolumn{1}{l}{68.00}  \\ 
\midrule  
\multirow{2}{*}{\textbf{SeaLLM}} &
  5-shots &
  \multicolumn{1}{l|}{75.80} &
  \multicolumn{1}{l|}{74.40} &
  \multicolumn{1}{l|}{69.60} &
  \multicolumn{1}{l|}{77.80} &
  \multicolumn{1}{l|}{72.00} &
  \multicolumn{1}{l}{69.60}  \\ 
 &
  0-shot &
  \multicolumn{1}{l|}{59.80} &
  \multicolumn{1}{l|}{53.20} &
  \multicolumn{1}{l|}{51.20} &
  \multicolumn{1}{l|}{60.60} &
  \multicolumn{1}{l|}{69.80} &
  \multicolumn{1}{l}{66.0}  \\ 
\midrule
\multirow{2}{*}{\textbf{Sailor}} &
  5-shots &
  \multicolumn{1}{l|}{60.80} &
  \multicolumn{1}{l|}{65.00} &
  \multicolumn{1}{l|}{58.40} &
  \multicolumn{1}{l|}{55.60} &
  \multicolumn{1}{l|}{75.40} &
  \multicolumn{1}{l}{76.60}  \\ 
 &
  0-shot &
  \multicolumn{1}{l|}{54.00} &
  \multicolumn{1}{l|}{52.80} &
  \multicolumn{1}{l|}{55.20} &
  \multicolumn{1}{l|}{53.20} &
  \multicolumn{1}{l|}{76.60} &
  \multicolumn{1}{l}{73.40}  \\ 
 \bottomrule
\end{tabular}
\vspace{-0.2cm}
\caption{XCOPA experiments results }
\label{tab: xcopa_ref_app}
}

\end{table*}

\section{Data Creation \& Review Instruction}

This section provides the guidelines for data creation and review.

\subsection{Data Creation}

You are tasked to create at least 110 instances of commonsense causal reasoning data in the COPA format. Two examples of COPA instances:

\vspace{0.2cm}

\hrule

\vspace{0.2cm}

\noindent \textbf{Type:} Cause. \\
\noindent \textbf{Premise:} Water flows from faucet.\\ 
\noindent \textbf{Correct Choice:} He turns the faucet open.\\
\noindent \textbf{Incorrect Choice:} The faucet is broken.

\vspace{0.2cm}

\hrule

\vspace{0.2cm}

\noindent \textbf{Type:} Effect. \\
\noindent \textbf{Premise:} Two cars crash into each other.\\ 
\noindent \textbf{Correct Choice:} I am stuck in traffic.\\
\noindent \textbf{Incorrect Choice:} There are 1000 victims.

\vspace{0.2cm}

\hrule

\vspace{0.2cm}

\noindent As you can see, we have a premise followed by two choices, with one being more plausible (hence correct) than the other. Note the cause and effect types. In the former, the choice causes the premise, and vice versa. As mentioned above, in this task, you should create data following the above format but injected with Indonesian local or cultural information. Two examples:

\vspace{0.2cm}

\hrule

\vspace{0.2cm}

\noindent (Cause) I got a mild stomachache. \textbf{I just ate seblak}. I just ate nagasari.

\vspace{0.1cm}

\noindent (Effect) That kid's got circumcised. \textbf{He received a lot of money}. His family booked a holiday trip.

\vspace{0.1cm}

\noindent (Cause) A guy shouted ``You're a dog!'' \textbf{His friend is annoying}. His friend is cute and sweet.

\vspace{0.1cm}

\noindent(\textbf{Bold} indicates the correct alternative.)

\vspace{0.2cm}

\hrule

\vspace{0.2cm}

\noindent For the first example, ``Seblak'' is a name of typically spicy food that is more likely to cause stomachache, rather than ``Nagasari'' which is a sweet dessert. For the second example, it is a common practice to gift money to a young boy who has just got a circumcision. For the third example, ``anjing'' (EN: ``dog'') is a common swear word in Indonesia. As you can see, in these three examples, it is impossible for a person who does not know Indonesian culture to know the correct answer, and your task is to create at least 110 such instances, using the standard formal Indonesian language.

You can also see that these three examples each showcase a different category of local information. For the seblak vs. nagasari example, the name or \textbf{terminology} of these foods is the source of locality. For the circumcision example, the \textbf{cultural context} surrounding circumcision itself is the source of locality. In the last example (dog), the \textbf{language} usage is the source of locality. Later, you will be tasked to acutely categorize your data into these three categories. But for now, to ensure the variety of your data, you should try to come up with roughly $\sim$50 terminologies, $\sim$50 cultural contexts, and $\sim$10 languages in your data. You should also ensure that the ratio between causes and effects of your data instances is 50:50.

During data creation, you should also take into account the following criteria:
\vspace{-0.25cm}
\begin{itemize}
    \setlength\itemsep{-0.3em}
    \item \textbf{Appropriateness}, ensure that your data contains the appropriate local or cultural nuance well-known by Indonesians, especially native Jakartans.
    \item \textbf{Difficulty}, ensure that your data is not too easy, but also not too difficult or obscure. Especially when making the incorrect choice, put something that is obviously incorrect for natives, but difficult to guess for foreigners.
    \item \textbf{Correctness}, ensure that the logical reasoning contained in your data is correct.
    \item \textbf{Ambiguity}, ensure that there is no ambiguity in wordings and in the choices. Check again your incorrect option, it might be the case that it is still plausible given the premise.
    \item \textbf{Ethics}, ensure your data is not discriminatory towards any person or organization.
    \item \textbf{Clarity and format}, check your capitalization, grammar, and spelling.
\end{itemize}
\vspace{-0.25cm}

\subsubsection{Data Peer Review}

You will be tasked to blindly review other creators' data. The two choices order will be randomly swapped so you cannot see the intended correct answer. Your task is to first try to pick one choice that you deem to be more plausible. Second, you will be asked to put a qualitative comment on each data if you feel that there is a problem with regard to appropriateness, difficulty, correctness, ambiguity, ethics, or clarity. You should also put a comment for any instance that you find to be duplicated, whether with your own data or with other data you have reviewed until now.

Once the peer review is done, all data creators will meet together to discuss data that has incorrect answers by the reviewers and data that has a qualitative review. The discussion will result in a decision on whether each data would be accepted, rejected, or slightly modified. Once this process is done, you should rework your own data again, make the required changes, and repeat these review processes all over again until at least 550 data are accepted.

\subsection{Data Categorization}

Once your data passes peer review, you are tasked to categorize your data. As briefly touched above, your data should be put into three categories.
\vspace{-0.25cm}
\begin{enumerate}
    \setlength\itemsep{-0.3em}
    \item \textbf{Language}. If your data uses non-literal words or phrases, then it falls into this category.
    \item \textbf{Local terminology}. If your data uses any Indonesian entities, famous people, food names, location names, abbreviations, local concepts, etc., then it falls into this category. Note that this means entities that originate from outside Indonesia, such as Pizza Hut, KFC, NATO, are not considered local terminologies.
    \item \textbf{Culture.} Any cultural context that does not pertain directly to the use of local terminologies or language goes into this category. If the cultural context arises from the local term or language \textit{immediately}, then it does not fall into this category.
\end{enumerate}
\vspace{-0.25cm}
Note that a single data instance should always fall into one or more categories. 
You may find that categorizing your data is not easy, as some things can be ambiguous. Here are some tips.

First, to ensure whether a local term is really local, you should try to google it first. If there is no direct one-to-one translation to your term, then you can be sure that it is indeed local. For instance ``kobokan'' is local because its closest English term, ``finger bowl'' does not mean exactly the same thing.

Second, you might find that, rather confusingly, some language usage can also be a local term. For example, ``polisi tidur'' (EN: ``speedbump'', literally ``sleeping police'') is a non-literal phrase, but it is also a local term because there is no one-to-one literal replacement in Indonesian for it. On the other hand, ``datang bulan'' is a language, but is not a local term because ``haid'' and ``menstruasi'' (EN: ``menstruation'') are appropriate literal replacements.

Third, you may find your data that uses local term is also cultural at the same time. For example, you may want to connect ``Lebaran Haji'' with animal slaughter. However, this should not be labeled as a culture because animal slaughter is implied immediately by ``Lebaran Haji''. On the other hand, you may want to connect ``Lebaran'' with the emptiness of Jakarta streets. This is OK to be labeled as culture because the cultural impact is not immediately implied.

\subsubsection{Categorization Peer Review}

In this step, you are tasked to review another data creator's categorization. You are asked to highlight any categorization that you are unsure/disagree with. Meanwhile, another creator will review your work in the same way. Any disagreement should be resolved together with your peer until a final categorization is achieved.

\subsection{Colloquial Form Translation}

Next, you are tasked to translate all your data which is in standard Indonesia, into their colloquial forms. To do this, you should imagine your data being spoken/talked in a day-to-day Jakartan conversational context, then within that context, transform your data in a natural way. For example, from a standard sentence ``Saya sedang dalam perjalanan ke sana.'' (En: ``I am on my way there.''), you can imagine a context where you are texting your close friends that you are on your way in a colloquial manner: ``Gw otw'', which is natural. You should not translate it to ``I sedang going ke there.'', which is non-standard but is highly unnatural.

To preserve variety in the colloquial forms, we will not be providing a detailed guideline here. You should use your own knowledge and experience of colloquial Indonesian and not seek outside influence too much. 

\subsubsection{Colloquial Peer Review}

In this step, you are tasked to review another creator's colloquial translation. You are asked to highlight any translation that you deem unnatural or inaccurate with respect to the original standard data.

Similarly, another creator will also review your work. Any disagreement should be resolved together with your peer until a final colloquial translation for each data is achieved.

\section{Human Scoring Guideline}

The following contains the elaborated guidelines for human scoring of \dataname. The original guideline is in Indonesian but we translate it to English for this appendix section.

\subsection{What is COPA?}
COPA stands for Choice of Plausible Alternatives. As the name suggests, COPA is a test that consists of a set of multiple-choice questions. Each question contains a premise or situation that serves as the basis for the question. The premise is followed by two alternative options. The participant's task is to choose the option that is most plausible among the two. Below are some example of COPA questions.

\vspace{0.2cm}

\hrule

\vspace{0.2cm}

\noindent \textbf{Example 1}\\
\noindent \textbf{Premise:} My little brother wakes up late.\\
\noindent \textbf{Option 1:} Mother is angry because it's Monday.\\
\noindent \textbf{Option 2:} Mother is angry because it's Sunday.

\vspace{0.2cm}

\hrule

\vspace{0.2cm}

\noindent The correct answer is Option 1. Although it is possible that the mother scolds the little brother for waking up late on Sunday, this situation is less likely compared to Option 1, which is more common because Monday is a school day.

\vspace{0.2cm}

\hrule

\vspace{0.2cm}

\noindent \textbf{Example 2}\\
\noindent \textbf{Premise:} The national team wins the Thomas Cup.\\
\noindent \textbf{Option 1:} My brother buys beer at the minimarket to celebrate.\\
\noindent \textbf{Option 2:} My brother buys pizza hut to celebrate.

\vspace{0.2cm}

\hrule

\vspace{0.2cm}

\noindent The correct answer in the context of Indonesian society is Option 2. While it is possible that the brother buys beer for the celebration, this is less common, and most minimarkets in Indonesia are prohibited from selling beer.

\vspace{0.2cm}

\hrule

\vspace{0.2cm}

\noindent \textbf{Example 3} \\
\noindent \textbf{Premise:} My little brother is wearing a uniform. \\
\noindent \textbf{Option 1:} He is going to school. \\
\noindent \textbf{Option 2:} He is going to play.

\vspace{0.2cm}

\hrule

\vspace{0.2cm}

\noindent The correct answer is Option 1.

\subsection{Effect vs Cause}
Participants will encounter two types of questions in COPA: Effect and Cause. Examples 1 and 2 are Effect-type questions, while Example 3 is a Cause-type question. Participants can connect the premise and options using phrases like "as a result" or "because" to confirm.

\subsection{Choosing the Most Plausible Option}
If participants find a question where both options seem equally plausible, they are still asked to choose the option that is more likely or more plausible. Factors that can be used as a basis for choosing include:
\vspace{-0.25cm}
\begin{enumerate}
    \setlength\itemsep{-0.3em}
    \item \textbf{Choose the most likely option in the context of Indonesia and Jakarta.} Both options may be plausible in an international context, but participants are instructed to choose the one that is more likely in the national context of Indonesia and the city of Jakarta.
    \item \textbf{Facts and statistics.} Consider factual information and statistics. For example, in Example 2, the fact that most minimarkets statistically do not sell beer and that, statistically, Indonesians do not consume alcohol should guide the decision.
    \item \textbf{Sensible stereotypes.} Consider stereotypes that make sense. For instance, the stereotype that graduates from Islamic schools are usually more proficient in reciting religious texts than those from public schools may be a sensible guideline.
    \item \textbf{Cause and effect relationships.} Consider cause-and-effect relationships that make sense. For example, it is more likely that eating spicy food causes stomachache rather than causing a headache.
    \item \textbf{Personal knowledge/common sense.} Utilize personal knowledge or common sense, as well as insights from individuals within the participant's family or social circle. This can provide additional context and perspectives that may aid in making a more informed choice.
\end{enumerate}
\vspace{-0.25cm}
\subsection{Technical Instructions}

Participants are required to provide their Gmail addresses to the specified contact. The questions will be sent via Google Sheets, where each row contains one question (1 premise and 2 options). Participants are instructed to choose the most plausible option by checking the checkbox provided next to each option.

The total number of questions for each participant is 559.  Participants are requested to complete all 559 questions on their own and refrain from consulting with third parties, including search engines and AIs such as ChatGPT. 

\subsection{Submission and Honorarium}
Participants are required to complete all questions by Sunday, September 17, 2023. Once finished, participants can notify the specified contact that the task is complete. After that, edit access for the participant on the provided Google Sheet will be revoked.

Participants will receive a basic honorarium of IDR 250,000 plus IDR 180 for each correct answer. For example, if all answers submitted are incorrect, the honorarium is IDR 250,000. If all answers are correct, the honorarium is IDR 250,000 + (180 x 559) = IDR 350,620.\footnote{Jakarta minimum wage is, as of December 2023, slightly below IDR 5,000,000 per month. Assuming 20 working days and 8 working hours per day, this translates to IDR 31,250 per hour. Each participant requires 2-4 hours to complete all questions, which means that we have paid at least double the minimum wage stipulated by the government.}

The honorarium will be given to participants at most one day after the calculations are completed and after the participant's bank information is provided to the specified contact (whichever occurs last).
\end{document}